\documentclass[journal]{IEEEtran}

\ifCLASSINFOpdf
\else
\fi
\usepackage{microtype}
\usepackage{graphicx}
\usepackage{subfigure}
\usepackage{booktabs} 

\usepackage{algorithmic}
\usepackage[ruled,vlined]{algorithm2e}

\usepackage{hyperref}
\usepackage{url}
\usepackage{amsfonts}       
\usepackage{nicefrac}       
\usepackage{amsmath}
\usepackage{xcolor}
\usepackage{dcolumn}
\usepackage{siunitx,booktabs}
\usepackage{enumitem}
\usepackage{float}
\usepackage{caption}
\usepackage{siunitx}
\usepackage{stackengine} 
\newcommand\oast{\stackMath\mathbin{\stackinset{c}{0ex}{c}{0ex}{\ast}{\bigcirc}}}

\makeatletter
\usepackage{tikz}
\newcommand*\circled[2][1.6]{\tikz[baseline=(char.base)]{
    \node[shape=circle, draw, inner sep=1pt, 
        minimum height={\f@size*#1},] (char) {\vphantom{WAH1g}#2};}}
\makeatother

\usepackage{hyperref}

\newcommand{\mask}{\boldsymbol{M}_{\boldsymbol{\Phi^{(i)}}}}

\newcommand{\maskr}{\boldsymbol{\widetilde{M}}_{\boldsymbol{\Phi^{(i)}}}}
\newcommand{\maskrL}{\boldsymbol{\widetilde{M}}_{\boldsymbol{\Phi}}}
\newcommand{\logits}{\boldsymbol{\Phi}}
\newcommand{\probs}{\boldsymbol{\pi}}
\newcommand{\Wmat}{\boldsymbol{W}^{(i)}}
\newcommand{\layeri}{g^{(i)}_{\boldsymbol{W}}}

\begin{document}

\title{Dynamic Probabilistic Pruning: A general framework for hardware-constrained pruning at different granularities}

\author{Lizeth~Gonzalez-Carabarin,
        Iris A.M.~Huijben,
        Bastiaan S.~Veeling, 
        Alexandre~Schmid
        and~Ruud J.G.~van Sloun}

\markboth{Journal of \LaTeX\ Class Files,~Vol.~, No.~,}%
{Shell \MakeLowercase{\textit{et al.}}: Bare Demo of IEEEtran.cls for IEEE Journals}

\maketitle

\begin{abstract}
Unstructured neural network pruning algorithms have achieved impressive compression rates. However, the resulting - typically irregular - sparse matrices hamper efficient hardware implementations, leading to additional memory usage and complex control logic that diminishes the benefits of unstructured pruning. This has spurred structured coarse-grained pruning solutions that prune entire filters or even layers, enabling efficient implementation at the expense of reduced flexibility.  Here we propose a flexible new pruning mechanism that facilitates pruning at different granularities (weights, kernels, filters/feature maps), while retaining efficient memory-organization  (e.g. pruning exactly k-out-of-n weights for every output neuron, or pruning exactly k-out-of-n kernels for every feature map). We refer to this algorithm as Dynamic Probabilistic Pruning (DPP). DPP leverages the Gumbel-softmax relaxation for differentiable k-out-of-n sampling, facilitating end-to-end optimization.
We show that DPP achieves competitive compression rates and classification accuracy when pruning common deep learning models trained on different benchmark datasets for image classification. Relevantly, the non-magnitude-based nature of DPP allows for joint optimization of pruning and weight quantization in order to even further compress the network, which we show as well. Finally, we propose novel information theoretic metrics that show the confidence and pruning diversity of pruning masks within a layer.
\end{abstract}


\IEEEpeerreviewmaketitle

\section{Introduction}

\begin{figure}[h]
        \centering

        \begin{subfigure}{(a)}
                \includegraphics[scale=0.32]{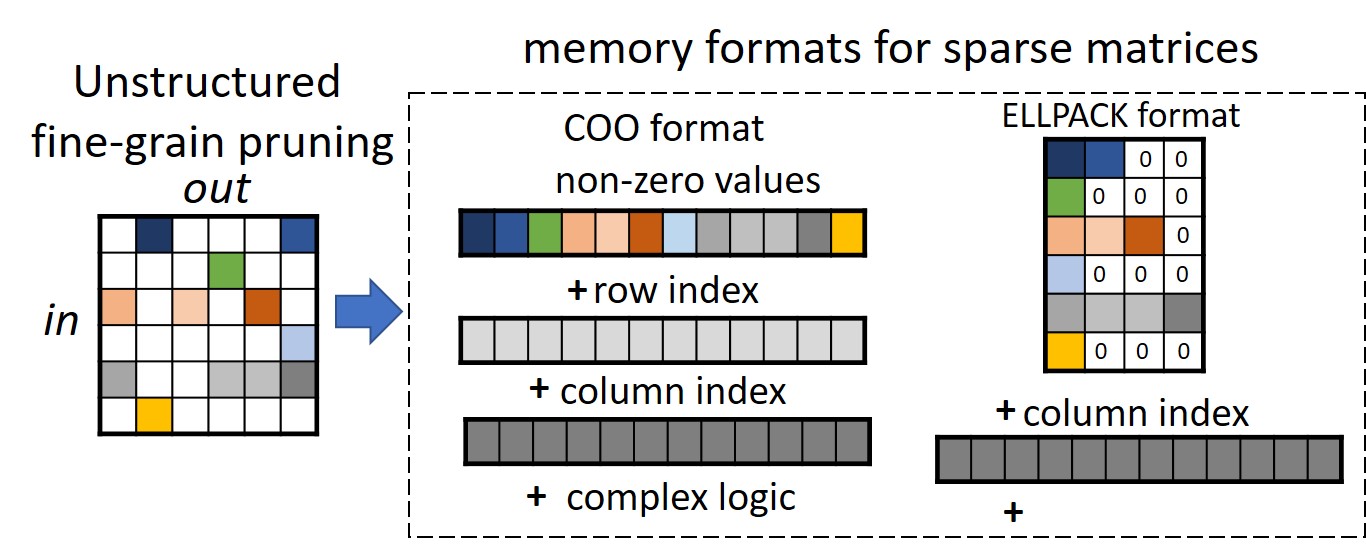}
                \label{fig:fig1a}
        \end{subfigure}  
        \newline
        \newline
        \begin{subfigure}{(b)}
          \includegraphics[scale=0.32]{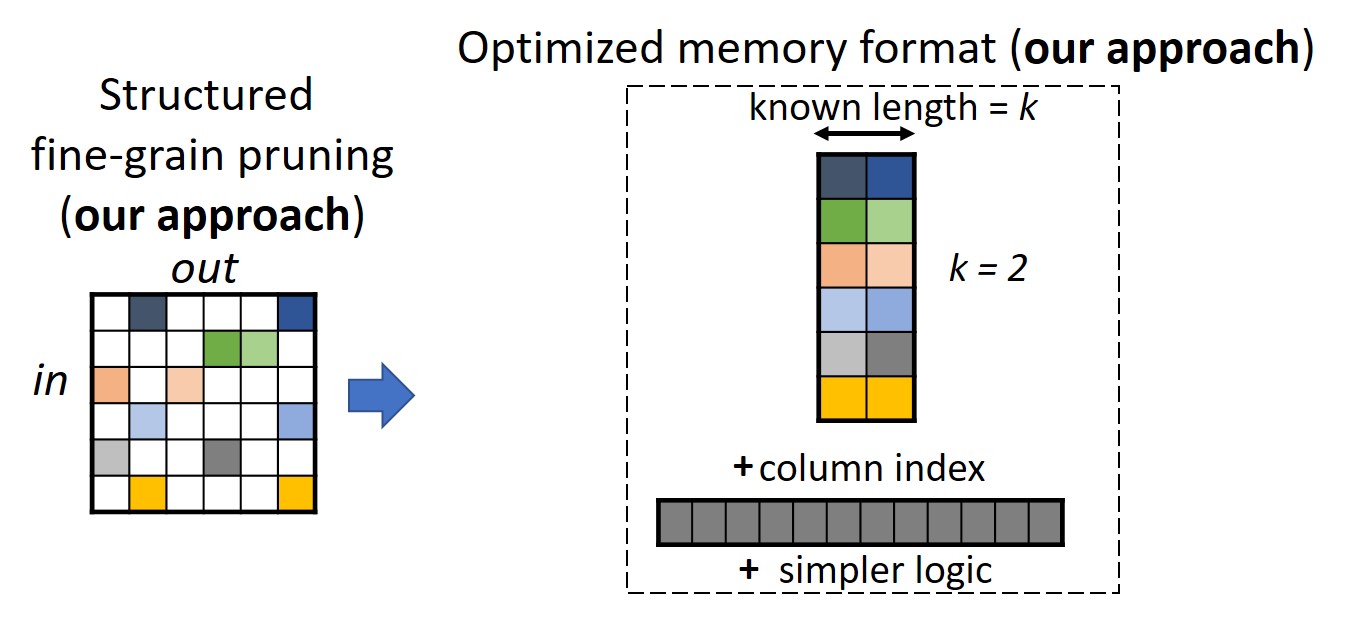}
                \label{fig:fig1b}
        \end{subfigure}
        \newline
        \caption{ Illustration of memory sparse formats of custom hardware for (a) unstructured fine-grained and (b) structured fine-grained pruning. DPP learns sparse patterns that follow a regular structure, facilitating memory organization by preventing zero-padding, and therewith allowing efficient bandwidth usage. }\label{fig:fig1}
\end{figure}

\IEEEPARstart{T}{he} evident success of Deep Learning (DL) models is accompanied by a steadfast growth in the number of hyperparameters and computational cost. This has become a bottleneck for hardware deployment, which is constrained to certain computational and memory budgets. For instance, the VGG-16 architecture occupies more than 500 MB of storage and performs 1.6 x $10^{10}$ floating-point arithmetic operations \cite{cheng_survey_2019,sze_hardware_2017}. In contrast, field-programmable gate array (FPGA)-based platforms are constrained to a few thousands computing operations, making them unsuitable for deployment of large DL models.

To shrink such big models, different solutions have been proposed such as quantization \cite{hubara2016} (and in the extreme case binarization \cite{courbariaux2016}), knowledge distillation, \cite{hinton2015}, weight sharing \cite{han2016}, and model pruning.

Pruning has gained notable attention since pruned model performance was found to yield on par performance compared with non-pruned counterparts. Remarkably, pruning has been shown to prevent overfitting as well \cite{lecun1990, hanson1989}.
Pruning algorithms can result in either structured or unstructured pruning strategies. Structured pruning methods remove model parts at the level of e.g. layer, channels, or individual filters \cite{liu2018_structured}. We refer to this kind of pruning as structured coarse-grained pruning. Unstructured pruning, on the other hand, prunes individual weights (i.e. it sets a fraction of weights to zero). We refer to this type of pruning as unstructured fine-grained pruning. Unstructured fine-grained pruning has achieved impressive compression rates, nevertheless implementing fine-grained sparse matrices in dedicated hardware is a challenging task due to the typically irregular distribution of non-zero values. Usually, to avoid storage of a large number of zeros, non-zero values are stored in  specific formats \cite{Dorrance14}. Several other works have proposed more sophisticated compressing coding techniques for sparse weights targeting hardware implementations  \cite{Fowers2014, Fujiki2019,Kreutzer2012}. For instance, COO format stores the indexes of columns and rows of non-zero values, and the ELLPACK format stores only the column indexes but a zero padding is required, wasting a large amount of memory and leading to poor bandwidth usage  (see Fig.\ref{fig:fig1}a). Besides, additional control logic is required to compute operations (e.g matrix multiplication) with such formats, increasing the complexity and power consumption for embedded applications \cite{Lu19_FPGA, Dey18,Yue19}. Therefore, we propose a framework that naturally generates structured sparsity for several levels of granularity, by fixing the number of active elements within a candidate set (comprising e.g. weights, kernels, filters) to $K$.

Fig.\ref{fig:fig1}b illustrates the benefits of our approach for unstructured fine-grained vs structured fine-grained pruning (our approach). The additional degree of freedom on selecting the granularity level allows to chose the best pattern depending on the network's architecture and/or application.  In addition, the framework is capable of integrating both pruning and quantization, to benefit from both techniques. The proposed method exploits the notion of Deep Probabilistic Subsampling (DPS) \cite{Huijben2020Deep} to dynamically generate stochastic pruning masks during training, which are independent of the magnitude of the weights, allowing for direct applications to quantize and even binarize weights in one run without the need of post-training finetuning. The main contributions of this work are the following:

\begin{itemize}

\item We propose Deep Probabilistic Pruning (DPP), which learns to generate hardware-oriented sparse structures for different levels of granularity: fine-grained (weights), medium-grained (kernels) and coarse-grained (filters), facilitating memory allocation and access for hardware implementations. We adopt a layer-wise sparsity level that can be selected by the user, which is beneficial in case of hardware constraints that dictate a maximum memory usage and specific patterns.

\item Thanks to the fact that DPP is not a magnitude-based pruning algorithm, it allows for joint optimization of pruning masks and quantization (and even binarization) of network parameters, producing ultra-compressed models with low-memory and low-complexity, suitable for further dedicated hardware.

\item Leveraging the probabilistic nature of DPP, we propose novel information-theoretic metrics that capture the confidence and diversity of the pruning masks leveraged within a network layer. We show how these metrics differ during training and between fully-connected and convolutional layers.
\end{itemize}

\section{Related work}
\label{related_work}

In this section we will give an overview on different pruning strategies. Network pruning approaches can be coarsely divided into single- and multi-stage strategies, strategies that prune at different granularities, i.e. fine-grained (weights), medium-grained (kernels), and coarse-grained (filters), and approaches that do or do not result in hardware-friendly pruned models. 

Hardware-friendly pruning is achieved when structure is present in the pruning pattern, since an unstructured selection results in memory storage using matrices that need (memory-inefficient) zero-filling. When pruning entire filters (or even layers), such structure is present, and therefore, conventionally, the term coarse-grained pruning is often used analogously with the term structured pruning. Similarly, fine-grained pruning is often referred to as unstructured pruning. Following the terminology introduced by \cite{zhou2021learning}, we slightly redefine terms and use \emph{structured pruning} for methods that result in \emph{hardware-friendly} pruned models at \emph{any granularity}. 

Our approach generalizes advantages of different methods into one general framework that - opposed to other approaches - facilitates \emph{single-stage}, \emph{hardware-friendly pruning} at \emph{any granularity}; fine-, medium- and coarse-grained. Additionally, it allows for joint optimization of pruning and quantization (and even binarization).

\subsection{Unstructured fine-grained pruning} 
The early work of \cite{han2015} proposed a three-stage pipeline for unstructured fine-grained pruning, which was later extended to deep compression \cite{han2016}. These works were followed by single-stage pruning approaches, that jointly/dynamically optimize both the model parameters and pruning process. The authors of \cite{zhu2017, narang2017} e.g. prune weights based on the gradual increment of sparsity during training, and a magnitude-based pruning framework in which a minimal sparsity value is set, is proposed at \cite{bellec2018} propose. 

Further the work of \cite{mostafa2019} proposes Dynamic Sparse Reparameterization (DSR) based on an adaptive threshold for pruning and an automatic reallocation of
pruning parameters across layers. The authors of \cite{dettmers2019} propose to select the weights with highest momentum, which significantly improved accuracy. All these works yield unstructured fine-grained pruning matrices, which lead to inefficiencies in terms of memory access and allocation in current hardware platforms \cite{ZhangCh15_FPGA,ZhangJ17_FPGA}.

\subsection{Structured coarse-grained pruning} To prevent unstructured pruning masks, previous work has explored pruning at the architectural level (rather than the weight level) such as pruning filters or layers. In fact, it has been experimentally shown that rather than eliminating weight connections, pruning at the architectural level may offer more benefits to reduce memory, while retaining state-of-the-art accuracy \cite{liu2018_structured}. Most works adopt a multi-stage approach, where pruning and model training happens disjointly \cite{He2017channel_prune,li_pruning_2017,Liu2017lecnn,wen_learning_2016,Luo2017thinet,dong2019network}. Such multi-stage methods typically suffer from performance drops after pruning, requiring fine-tuning steps. The magnitudes of the weights are often used as indicator whether or not to prune an entire structure, \cite{li_pruning_2017} e.g. prunes filters based on the sum of the absolute magnitudes of the corresponding weights, and \cite{wen_structured_2016} proposed the Structured Sparsity Learning (SSL) method to prune filters, channels, and depth structures. Different from the magnitude-based approaches, \cite{Hu_structured_2016} rely on the output of activation layers and calculate an average percentage of zeros as a weighting for the filter relevance. The authors of \cite{kang2020operationaware, Wang2020DynamicNP} leverage, opposed to the other methods, a single-stage approach. The work of \cite{Wang2020DynamicNP} most resembles our generalized framework, when applied on the coarse level. However, we are guaranteed to exactly select k-out-of-n filters, while this number k is only approximated using a proxy for the $\ell_0$ penalty during training in the work of \cite{Wang2020DynamicNP}. To conclude, all methods reviewed in this paragraph prune at coarse-grained level (layers or filters), and have not been transferred to prune models at fine-grained levels (i.e. weights), making them more restricted than our generalized framework that is suitable for pruning any of the granularities.

\subsection{Structured fine-grained pruning}
To the best of our knowledge only few works have addressed the issue of structured pruning of fine-grained elements (weights). 
The authors of \cite{Ma19} introduce convolutional sparse patterns for hardware-oriented pruning, however this approach limits the possible pruning patterns to only a few options, reducing its flexibility. The work of \cite{li2020penni} also introduces a hardware-friendly sparse approach for kernel pruning, however it requires several stages to achieve competitive results, which contrasts with our single-stage approach. The authors of \cite{nvidia20} also proposed the generation of structured fine-grained pruning by generating sparse patterns compatible with their GPU A100. Their approach also considers a multi-stage approach, and it is focused on fine-grained pruning, without exploring other levels of granularity. Recently, \cite{zhou2021learning} extended the work of \cite{nvidia20} to a single-stage approach, by jointly training the model to generate structured fine-grained pruning from scratch. Their method in its current form, however, is only applicable to fine-grained pruning, since the pruning mask is directly created from the magnitudes of the weights. Moreover, the authors adopt the straight-through estimator (STE) to circumvent the non-differentiable pruning procedure, while we adopt a more principled gradient estimator, dedicated to differentiable subset sampling \cite{xie2019reparameterizable}.

\subsection{Pruning and quantization}
Works that optimize pruning and quantization offer ultra-compressed models for higher memory savings for unstructured pruning \cite{Shaokai2018} \cite{Tung2018}, and structured pruning \cite{Yang_He_Fan_2020,han2016,Tung2018,Wang2018haq,Shaokai2018,wang2020apq,Yang2019ann,Yang_He_Fan_2020,zhao2019}. Nevertheless, many of them require several stages during training to achieve this integration; none of these works has achieved joint optimization of quantization and structured fine-grained pruning.

 \section{Methodology}
\label{methodology}
\subsection{Notation}
\label{sec:notation}
We introduce a neural network with $L$ layers, indexed with $i$. Each layer is parameterised by a 

matrix $\boldsymbol{W}^{(i)} \in \mathbb{R}^{N^{(i-1)}\times a \times N^{(i)}}$,  and a bias vector $\mathbf{b}^{(i)} \in \mathbb{R}^{N^{(i)}}$ (which we ignore in the rest of the notations), where $N^{(i-1)}$ and $N^{(i)}$ are the number of feature maps (or channels) of layer $i-1$ and $i$, respectively, and $a~(= \sqrt{a} \times \sqrt{a})$ denotes the size of a 2D convolutional kernel.

The output of the $i^{\text{th}}$ layer can be defined as \mbox{$\boldsymbol{x}^{(i)} = \layeri(\boldsymbol{x}^{(i-1)})$}, where the functionality of $\layeri()$, depends on the layer being e.g. fully-connected or convolutional, and linearly or non-linearly activated.  Note that the fully-connected layer is a special case of a convolutional layer with $a = 1$.

\subsection{Dynamic masking based on probabilistic subsampling } \label{DPPgeneral}
\begin{figure}[htp]
    \centering
    \includegraphics[scale=0.35]{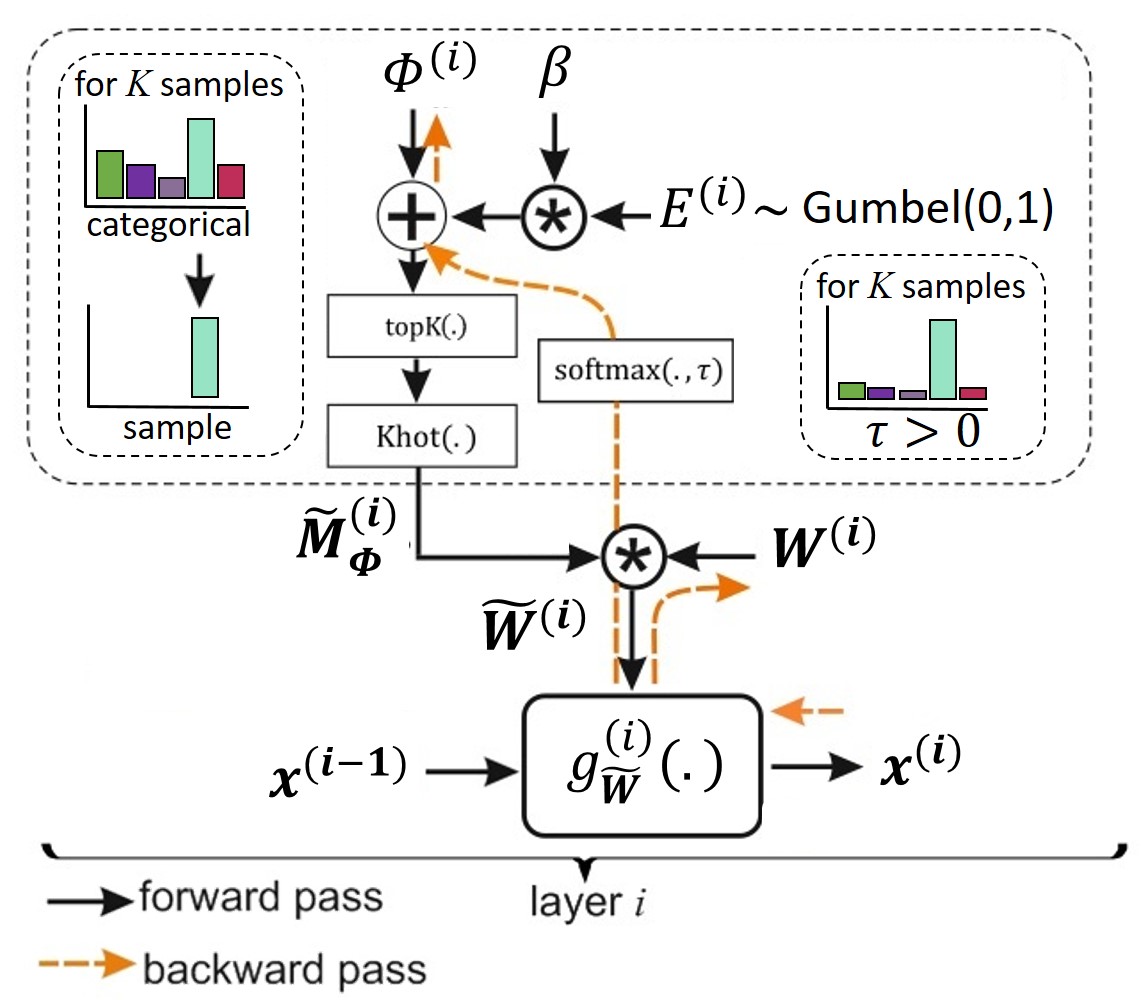}
    \caption{An illustration of deep probabilistic pruning (DPP) for dynamically masking of weights $\Wmat$ with a binary mask realization $\maskr$. This mask is generated by sampling from a categorical distribution with probabilities that are jointly trained with the model weights $\boldsymbol{W}$. The  symbol \circled[0.5]{$\star$} indicates element-wise multiplication.}
    \label{fig:Fig2}
\end{figure}

We aim to jointly optimize the parameters of the model while simultaneously learning to prune them. In this section we explain how our framework, which we refer to as Dynamic Probabilistic Pruning (DPP), achieves this joint learning. 

For all layers $\layeri()$, with $i \in \{1,\ldots,L\}$, we introduce a binary mask $\mask \in \{0,1\}$, parameterised by $\logits^{(i)}$. By means of element-wise multiplication it activates a subset of $S$ elements from $\Wmat$. Generation of these masks $\mask$ follows the DPS-topK framework \cite{Huijben2020Deep}, on which we will elaborate here.

The authors of \cite{Huijben2020Deep} propose an end-to-end framework for joint learning of a discrete sampling mask with a downstream task model by introducing DPS; a parameterised generative sampling model:
\begin{equation}
 P(\mask| \logits^{(i)}).
\end{equation}

Trainable parameters $\logits^{(i)} \in \mathbb{R}^{N^{(i-1)}\times a \times N^{(i)}}$ denote unnormalized log-probabilities (logits). In order to generate a binary sampling mask realization $\maskr$ from $\logits^{(i)}$, we select/sample exactly $K$ unique values over one axis of $\logits^{(i)}$. This implies that pruning takes place over this same axis in $\boldsymbol{W}^{(i)}$ (e.g. pruning the $N^{(i)}$ channels of layer $i$, implies sampling over the axis of length $N^{(i)}$). As this pruning axis differs per case in our experiment section, we denote it with \textit{p-axis} (pruning axis) for now. Each element in $\logits^{(i)}$ is thus the (unnormalized) log-probability of activating the corresponding element in $\Wmat$. Note that thanks to the fact that the binary mask $\maskr$ is parameterised on $\logits^{(i)}$ rather than $\Wmat$, we can combine DPP with quantization of the values in $\Wmat$, which we can jointly learn as well.

Similarly as in \cite{Huijben2020Deep,huijben2020_MRI}, we adopt Gumbel top-K sampling \cite{Gumbel1954StatisticalApplications, kool2019stochastic} to draw $K$ unique elements from the elements in the pruning axis. We denote this operation by $\mathrm{topK}_{\textit{p-axis}}(\cdot)$, where the subscript \textit{p-axis} indicates that sampling is only performed over the pruning axis. In order to create a binary sampling mask of equivalent size as $\mathbf{W}^{(i)}$, we transform the samples to a $K$-hot vector, which contains $K$ ones at the selected indices, and zeros at the remaining/non-selected positions.
Formally, we define the binary mask realization $\maskr$ as:
\begin{align}
    \label{eqn:MhotSample}
    \maskr
    &= \mathrm{Khot_{\textit{p-axis}}}\big\{\mathrm{topK_{\textit{p-axis}}}(\logits^{(i)}+\beta \mathbf{E}^{(i)})\big\},
\end{align}

where $\mathbf{E} \in \mathbb{R}^{N^{(i-1)}\times a \times N^{(i)}}$
are i.i.d. Gumbel noise samples from $\operatorname{Gumbel}(0,1)$, scaled with a scalar $0 < \beta \leq 1 $. Note that the pruning axis relates to only one of the three axes of the 3-dimensional matrix $\maskr$. Therefore, the total number of active elements $S$ in $\maskr$ does not equal $K$, but $K$ times the size of the two remaining axes. 

If we would allow $S$ ones to be distributed randomly within $\maskr$, it would result in unstructured pruning of the model when applying this mask element-wise on $\Wmat$. However, by demanding exactly $K$ elements to be active over the pruning axis, we enforce structure in the binary mask $\maskr$. To define structure at different granularities (e.g. pruning weights, kernels or filters), we `tie' together certain elements in $\logits^{(i)}$, such that they all update equivalently during training. It effectively reduces the number of trainable logits within $\logits^{(i)}$. As a result, if for example all logits over the kernel axis with size $\sqrt{a} \times \sqrt{a}$ are tied, a kernel will be either activated or deactivated as a whole, rather than on weight level. Section \ref{granularity} elaborates on structured pruning at different granularities. 

During backpropagation the $\mathrm{Khot}_{\textit{p-axis}}~\circ~\mathrm{topK}_{\textit{p-axis}}$ operation must be relaxed as it is non-differentiable. The work of \cite{huijben2020_MRI, Huijben2020Deep} proposes to adopt the Gumbel-softmax relaxation \cite{Jang2016CategoricalGumbel-Softmax, Maddison2017TheVariables} for that. It relaxes the non-differentiable $\operatorname{argmax}$ operation using a temperature ($\tau$)-parameterised $\mathrm{softmax}_{\tau}(\cdot)$  function. The authors of \cite{kool2019stochastic} showed that sampling $K$ times without replacement from the same distribution is equivalent to top-K sampling, and \cite{xie2019reparameterizable} showed that iterative sampling without replacement from its relaxed counterpart (using the softmax), is a valid top-K relaxation. As such we can directly leverage Gumbel-softmax sampling without replacement during backpropagation in order to flow gradients to $\logits^{(i)}~\forall i \in \{1,\ldots,L\}$. Figure \ref{fig:Fig2} and Algorithm 1 provide a schematic overview and pseudocode, respectively, of the full training procedure.

\begin{algorithm}[h!]
\caption{Dynamic Probabilistic Pruning (DPP)}
\label{DPPalgorithm}
\begin{algorithmic}
\REQUIRE  Training dataset $\mathcal{D}$, neural network with $L$ layers, and initialized trainable parameters $\{\logits\}$, $\boldsymbol{W}$, $\boldsymbol{b}$\}, Number of active elements $K$, Pruning axis \textit{p-axis}, Gumbel noise scaling $\beta$, temperature annealing settings $\{\tau_{\text{init}},\tau_{\text{end}} \} = \{5.0, 0.5\}$, Loss scaling $\mu$, number of epochs $n_{\text{iter}}$.
\ENSURE     
Model with trained parameters $\boldsymbol{W}$ and $\boldsymbol{b}$, binary mask realizations $\maskrL$ parameterised by $\logits$.

\STATE{ - Compute: $\Delta\tau = \frac{\tau_{\text{init}} - \tau_{\text{end}}}{n_{\text{iter}}-1}$ \\}

\FOR{$n=1$ to $n_{\text{iter}}$}
    \STATE{ // \textit{Forward pass}  }
    \STATE{ - Draw random batch $\boldsymbol{x}_n \sim \mathcal{D}$}
    \FOR{$i = 1$ to $L$}

           \STATE{- Draw i.i.d. Gumbel noise samples: $\mathbf{E}^{(i)}$}  \\
           \STATE{- Sample binary mask: \\ 
           $\maskr = \mathrm{Khot_{\textit{p-axis}}} \big\{\mathrm{topK_{\textit{p-axis}}}(\logits^{(i)}+\beta \mathbf{E}^{(i)})\big\} $ }  

        \STATE{- Apply mask: $\widetilde{\boldsymbol{W}}^{(i)} = \boldsymbol{W}^{(i)}\oast \maskr $}
    \ENDFOR\\
    
    \STATE{ - Compute output:~$\hat{\boldsymbol{x}}_n = g_{\widetilde{\boldsymbol{W}}}^L\circ g_{\widetilde{\boldsymbol{W}}}^1(\boldsymbol{x}_n)$\\
    \STATE{ - Compute loss: $\mathcal{L}_{CE}(\hat{\boldsymbol{x}}, \boldsymbol{x}) + \mathcal{L}_{e}(\logits)$}\\
    \STATE{ // \textit{Backward pass}  }
    \STATE{ - Set: $\tau = \tau_{\text{init}} - (i-1) \cdot \Delta \tau$} \\
    \STATE{ - $\nabla_{\logits} \widetilde{\boldsymbol{M}}_{\logits} \propto \nabla_{\logits} 
    \mathbb{E}_{\boldsymbol{E}}\big[
    \mathrm{softmax}_{\textit{p-axis}} \{ \frac{\logits+\beta\boldsymbol{E}}{\tau} \} \big]$}
    \STATE{- Update: $\{\logits, \boldsymbol{W}, \boldsymbol{b}\} \propto \mathcal{L}_{CE}(\hat{\boldsymbol{x}},\boldsymbol{x}) + \mu\mathcal{L}_{e}(\logits)$}
    }

\ENDFOR
\end{algorithmic}
\end{algorithm}

\subsection{DPP for pruning different levels of granularity}
\label{granularity}

As explained in the previous section, DPP learns a binary mask $\maskr$ that selects exactly $S$ elements from $\Wmat$. By connecting trainable log-probabilities in $\logits^{(i)}$ during training, we enforce pruning at different granularaties. We define three different pruning scenarios: \textbf{a)} Fine-grained pruning (DPP-F) \textbf{b)} Medium-grained pruning (DPP-M) and \textbf{c)} Coarse-grained pruning (DPP-C). 

\begin{figure}[h!]
        \centering
        \begin{subfigure}{\textbf{a. Fine-grained pruning (DPP-F):} pruning weights }
             \includegraphics[trim=0 1 0 10, scale=0.39]{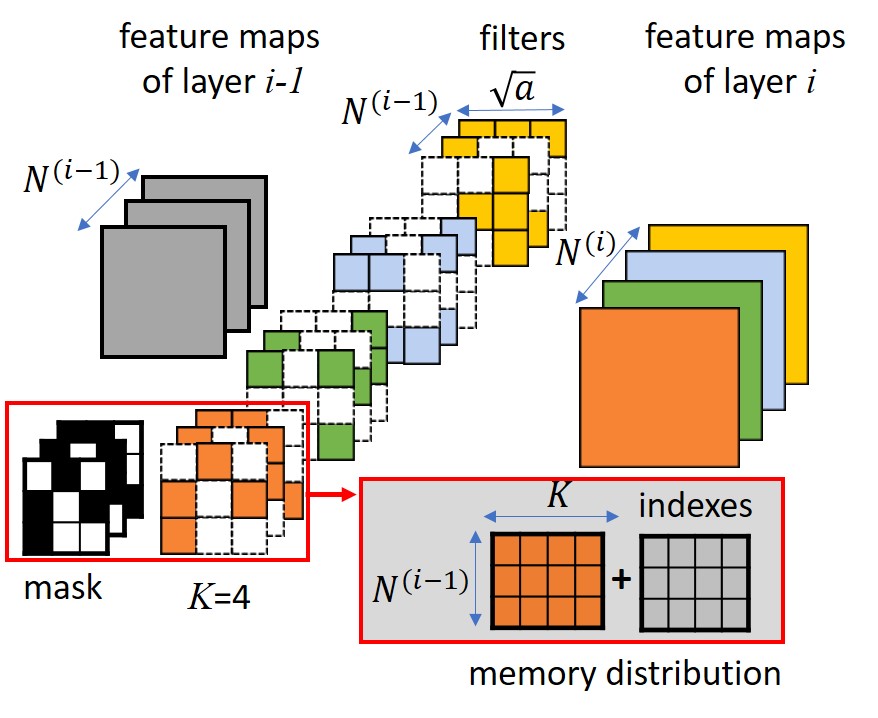}
                \label{fig:fig3a}
        \end{subfigure}  
        \newline
        \begin{subfigure}{\textbf{b. Medium-grained pruning (DPP-M):} pruning kernels }
          \includegraphics[trim=0 1 0 10, scale=0.39]{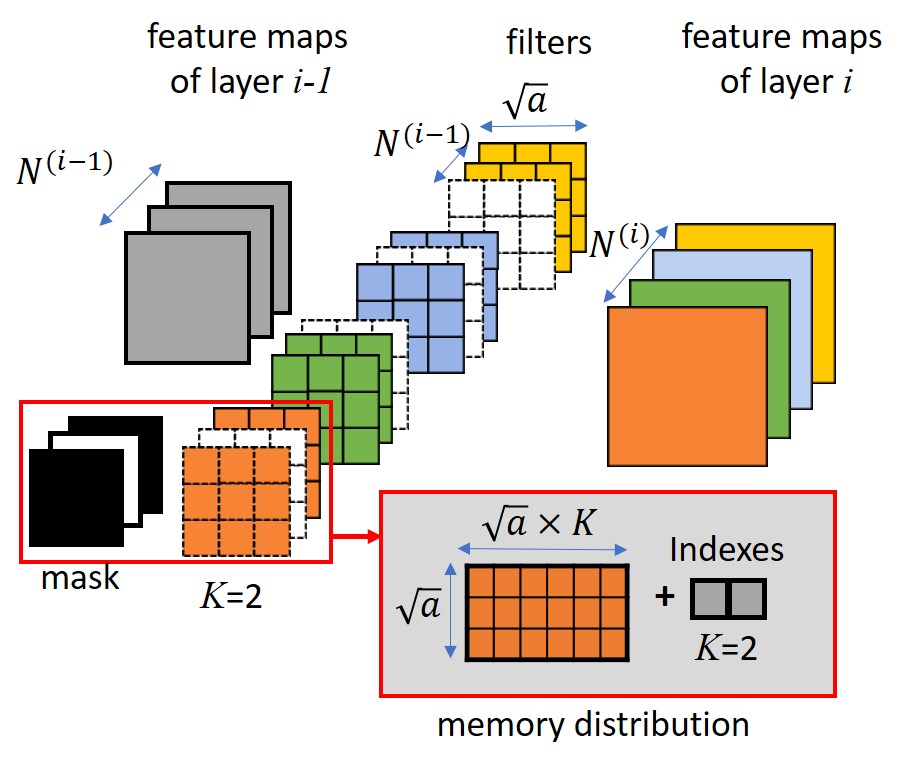}
                \label{fig:fig3b}
        \end{subfigure}
        \newline
        \begin{subfigure}{\textbf{c. Coarse-grained pruning (DPP-C):} pruning filters}
              \includegraphics[trim=0 3 0 10, scale=0.39]{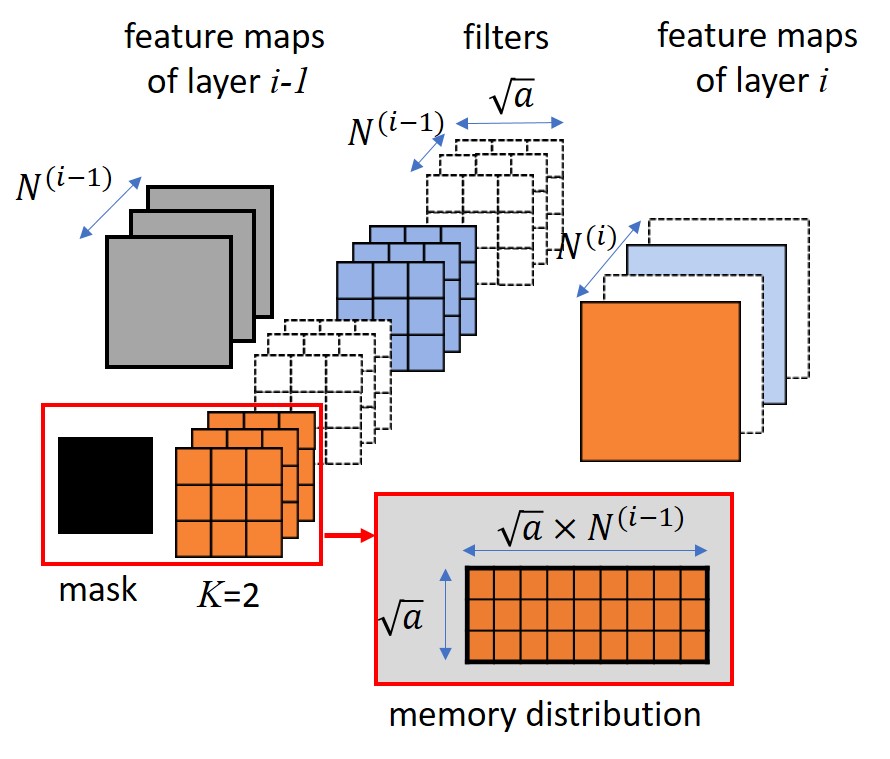}
                \label{fig:fig3c}
        \end{subfigure}

        \caption{Visualization for activating exactly $K$ elements at three different granularities (weights, kernels, filters). All adopted values are illustrative. Black squares of masks denote selected connections. We also show how the structure in our sparse matrices enables efficient memory implementation for all three cases.}
        \label{fig:fig3}
\end{figure}

Figure \ref{fig:fig3} illustrates the three scenarios. DPP-F activates $K$ out of $a$ kernel weights for each (2D) kernel within each (3D) filter.
DPP-M on the other hand activates $K$ (out of $N^{(i-1)}$) kernels per $N^{(i)}$ feature maps of layer $i$. Finally, DPP-C activates $K$ entire filters per layer. Table \ref{table1} summarizes the three different scenarios, and also indicates the number of values to be stored in memory in case of hardware implementation of the pruned model (e.g. on an FPGA). Additionally, for clarification we indicate the effective number of trainable logits within $\logits^{(i)}$ as a results of connecting logits over certain axes to enforce pruning at different granularities. Taking DPP-M as an example; as we prune $K$ out of $N^{(i-1)}$ entire kernels, the weights axis (of size $a$) has tied logits, as either all or none of the weights in a kernel are activated by the binary mask. We can interpret the resulting 2-dimensional logits matrix $\logits^{(i)}$ as containing the log-probabilities of $N^{(i)}$ number of categorical distributions, each containing $N^{(i-1)}$ classes. In order to generalize the three scenarios, we define $D$ as the number of such independent categorical distributions, and $C$ as the number of classes of each of these distributions. 

In the particular case of pruning connections within fully-connected layers, DPP activates $K$ (out of $N^{(i-1)}$) input neurons for each $N^{(i)}$ output neurons. Given the different granularity definitions we proposed, this pruning situation fits to DPP-M, with $a=1$. However, as the weights of the input neurons are the smallest possible entity to be pruned in fully-connected layers, this case is in literature often referred to as structured fine-grained pruning \cite{zhou2021learning, nvidia20}. In the rest of this paper, we therefore use DPP-F when referring to the special case of pruning connections in fully-connected layers.

\begin{table*}[t]
\caption{DPP for structured pruning for different granularity levels, and their corresponding pruned axis. For the estimation of the amount of stored values, the number of necessary indexes to point non-zero values plus the amount of non-zero elements $m$ is considered.}
\label{table1}
\vskip 0.15in
\begin{center}
\begin{small}
\begin{tabular}{cccccccc}
\toprule
 \textbf{Granularity}    & \textbf{Pruning axis}     & \textbf{Effective nr of trainable} &  D: \textbf{nr of independent} & S: \textbf{nr of active}  & \textbf{Stored values}\\ \textbf{level}
        &   ($K$ out of $C$)  & \textbf{logits per layer i} & \textbf{categorical distr.} & \textbf{weights in} $\Wmat$ &   \textbf{per layer} $i$\\

\midrule
\midrule
 Fine    & Kernel weights  & $N^{(i-1)} \times a \times N^{(i)}$ & $N^{(i-1)} \times N^{(i)}$ & $N^{(i-1)} \times K \times N^{(i)}$ & $2S$ \\
(DPP-F)  &  ($K$ out of $a$)       &  &   \\
 \midrule
 Medium & Kernels  & $N^{(i-1)} \times 1 \times N^{(i)}$ & $1 \times N^{(i)}$ & $ K \times a \times N^{(i)} $ & $S + KN^{(i)}$\\
(DPP-M) & ($K$ out of $N^{(i-1)}$)        \\
 \midrule
Coarse  &  Filters  & $1 \times 1 \times N^{(i)}$ & $1 \times 1$ & $N^{(i-1)} \times a \times K$  & $S$\\
(DPP-C) &  ($K$ out of $N^{(i)}$)   \\

\bottomrule
\end{tabular}
\end{small}
\end{center}
\vskip -0.1in
\end{table*}

\subsection{Training details}

Here we elaborate on training details of DPP. We jointly train the model parameters $\{\boldsymbol{W}, \boldsymbol{b}\}$ and the unnormalized logits in $\logits$, by means of error back-propagation of the total loss with respect to these parameters. The down-stream task model loss is defined as the cross-entropy between the targets and the predictions (denoted with $\mathcal{L}_{CE}$). Also, to encourage sparse distributions in $\logits$, as proposed by \cite{Huijben2020Deep} (eq. 8), we penalize these unnormalized log-probabilities with an entropy penalty $\mathcal{L}_{e}$, which we multiply by $\mu$. 
During training, the unnormalized log-probabilities $\logits^{(i)}$ ($\forall i \in L$) are constantly being updated. Also, the realization of the Gaussian noise matrix $\mathbf{E}^{(i)}$ differs per element within a mini-batch and over training/epochs. As a result, binary mask realizations vary during training and within the mini-batches, allowing the model to efficiently explore different pruned model instantiations. While in the original Gumbel-max trick \cite{Gumbel1954StatisticalApplications} the Gumbel noise is typically un-scaled when sampling from the distribution, heuristically, we found improved model performance when down-scaling the Gumbel noise with a factor $\beta$. In the experiments where we combine DPP with parameter quantization, we follow the quantization procedure proposed by \cite{courbariaux15_bc}. 
In all experiments we prune (and in some cases quantize) only $\boldsymbol{W}$, and not $\boldsymbol{b}$, as the weights in $\boldsymbol{W}$ contribute to the largest part of the parameters in the model. 

During inference, one binary mask $\maskr$ per layer is drawn from the trained log-probabilities $\logits^{(i)}$, which is then used to prune the model and compute the performance on the test set. 

\subsection{Information-theoretic metrics on sparsity confidence and diversity}
\label{sec:metrics}
To get insight into the training dynamics of DPP, we are interested in the change of confidence and diversity of the pruning patterns as training progresses. The probabilistic nature of DPP enables the use of the information theoretic measures, entropy and mutual information, to evaluate this. We compute these metrics per layer $i$ due to the heterogeneity of the masks between layers.

As defined in Section \ref{granularity}, within each layer $i$, $D~(\geq 1)$ number of independent categorical distributions, each with $C$ classes are being trained, from which we sample $K$ out of $C$ elements (weights, kernels or filters) without replacement. The average entropy of these $D$ pruning distributions tells us how confident the sparsity patterns on average are within this layer; the lower this $\mathrm{Average~Pruning~Entropy}$, the sparser the distributions, and thus the more certain the model is about the binary mask to be applied. Note, this metric can only be computed for DPP-F and DPP-M, as DPP-C implies $D=1$ (see Table \ref{table1}).

We can measure the average entropy from these $D$ pruning distributions in the $i^{\text{th}}$-layer mask marginal probabilities $\lbrace\probs^{(i)} \in \mathbb{R}^{D \times C}: 0 \le \pi^{(i)}_{d,c} \le 1, \sum_c \pi^{(i)}_{d,c} = K\rbrace$. No tractable function exists to compute marginal probabilities $\probs$ from the unnormalized log-probabilities in $\logits$. Instead we can easily take a Monte Carlo estimate by computing the average of $T$ realizations of $\mask$: 

\begin{align}
\probs^{(i)} \approx \dfrac{1}{T}\sum^T_{t=1} \maskr, ~~~\maskr \sim P(\mask| \logits^{(i)}),
\end{align}

which can effectively be estimated in parallel after every epoch for $T=100$. It's trivial to show that the entropy of any Gumbel-top-K distributed variable $x$ can be computed using the typical Shannon entropy \mbox{$H(x) = -\sum^C_{c=1} P(c\in x) \log P(c \in x)$}, and is upper bounded by $-K\log(K/C)$. As such, we can compute the $\mathrm{Average~Pruning~Entropy} $ for layer $i$: 
\begin{align}
    H(\mask|d) &= \dfrac{1}{D} \sum^{D}_{d=1} [- 
     \sum^{C}_{c=1} \probs^{(i)}_{d, c}\log \probs^{(i)}_{d, c}].
\end{align}
Furthermore, we can measure the diversity of the different sparsity patterns (within layer $i$), that result from sampling from the $D$ independent categorical distributions. This diversity can be measured as the mutual information between the different masks in layer $i$.

This $\mathrm{Pruning~Diversity}$ metric can be formalized as:
\begin{align}
 I(\mask,d) &= H(\mask) - H(\mask|d),
\end{align}

where $H(\mask)$ denotes the entropy of the average mask in layer $i$:
\begin{align}
    H(\mask) &= -\sum^{C}_{c=1} \widetilde{\pi}^{(i)}_{c} \log \widetilde{\pi}^{(i)}_{c}, \hspace{1cm} \mathrm{with} \nonumber \\
    \widetilde{\probs}^{(i)} &= \dfrac{1}{D} \sum^{D}_{d=1} \probs^{(i)}_{d} \hspace{0.2cm} \in \mathbb{R}^{1\times C}.
\end{align}

\section{Experiments}

We first assess DPP for pruning weights (FPP-F) on small convolutional and fully-connected architectures (LeNet) for the MNIST dataset. Additionally, we will demonstrate the performance in LeNet architectures in combination with quantized and binary weights based on \cite{courbariaux15_bc}. Across these experiments, we set a $K$ value per layer, which determines the exact number of non-pruned weights assigned to each output neuron. In the case of convolutional layers, $K$ is the number of non-pruned kernel weights per input feature map. Second, we test DPP for structured medium- and coarse-grained pruning (DPP-M and DPP-C) in medium-size convolutional networks (VGG-16 and MobileNet v1). It is important to notice that structured fine-grained pruning (DPP-F) is used for fully-connected layers.  For DPP-C, we use $K$ as a selector of the number of kernels or filters that must remain active in each layer, while for DPP-M, $K$ is the number of non-pruned kernels per output feature map. Additionally, we integrate quantization for medium-size datasets (VGG-16). We train all our models from scratch, without using any pre-trained model. For the results, non-pruned accuracy refers to the baseline accuracy for non-pruned networks, while pruned accuracy refers to the accuracy obtained after the network is pruned.  $\Delta_{acc}$ is obtained after the network is pruned with the remaining parameters. Since in some experiments, DPP is jointly integrated with quantization \cite{courbariaux15_bc}, the bit representation is presented, as well as the compression rate after pruning and quantization. For the compression rate, we consider the additional required memory to store the indexes of non-pruned values.

\subsection{MNIST}
We evaluate DPP first on MNIST benchmark dataset consisting of a total of 70,000 grayscale images of handwritten digits having a size of 28 $\times$ 28 pixels. We use 60,000 images for training and 10,000 images for testing. We evaluate the performance on 2 architectures. First, we use LeNet 300-100 \cite{lecun1998}, which consists of two fully-connected layers of 300 and 100 units, respectively. For this experiment, we use structured fine-grained pruning or DPP-F.    Second, we use LeNet-5 Caffe, which consists on two convolutional layers (20 and 50 filters respectively) followed by one fully-connected layer and a classification layer \cite{lecun1998}. Both, the convolutional and fully-connected layer are pruned with DPP-F.
We compare DPP-F with one of the latest fine-grained pruning algorithms \cite{liu2020} and show results in Table \ref{table:2}. 

\begin{table*}[h!] 
\caption{Experimental results of DPP for MNIST dataset using LeNet architectures  \label{table:2}}

\centering
\begin{small}
\begin{tabular}{@{}llllllll@{}}\toprule
\textbf{Network} & \textbf{Model}  & \textbf{Non-pruned} & \textbf{Pruned}  & \textbf{$\Delta_{acc}$}  & \textbf{Remain. (\%)} & \textbf{Bits} & \textbf{Compression} \\
\textbf{} & \textbf{}  & \textbf{acc.  (\%)} & \textbf{acc.  (\%)}  & \textbf{}  & \textbf{params.} & \textbf{} & \textbf{rate} \\\midrule
\textbf{}             &\textbf{\cite{liu2020}}   & 98.16 & 98.03 & \textbf{-0.13} & 2.48          & 32 & 13,44x\\ 
\textbf{}             &\textbf{DPP-F} (This work) & 98.19 & 97.90 & -0.29          & 1.95 & 32 & 25.64x\\ 
\textbf{LeNet300-100} &\textbf{DPP-F} (This work) & 98.19 & 97.82 & -0.37          & 6.71          & 8  & 29.80x\\
\textbf{}             &\textbf{DPP-F} (This work) & 98.19 & 96.81 & -1.38          & 21            & 2 & \textbf{38.09x}\\ \midrule

\textbf{}             &  \textbf{\cite{liu2020}}            & 99.18 & 99.11 & -0.07 & 1.64           & 32 & 20.32x\\
\textbf{}             &  \textbf{DPP-F} (This work) & 99.23	& 99.23	& \textbf{0.0}          &  2.5 & 32 & 20x\\ 
\textbf{LeNet5-Caffe} &  \textbf{DPP-F} (This work) & 99.23 & 99.00 & -0.23          & 4.1            & 8  & 48.78x\\
\textbf{}             &  \textbf{DPP-F} (This work) & 99.23	& 98.36	& -0.87          & 4.1            & 2  & \textbf{195.12x} \\

\bottomrule
\end{tabular}
\end{small}
\end{table*}

\begin{table*}[h!] 
\begin{minipage}{\textwidth}

\caption{Experimental results of DPP for CIFAR-10 and CIFAR-100 datasets using VGG-16 and MobileNet v1 architectures \label{table:3}}
\centering
\begin{small}

\begin{tabular}{@{}lllllllll@{}}\toprule
\textbf{Dataset} & \textbf{Network} & \textbf{Model}  & \textbf{Non-pruned} & \textbf{Pruned}  & \textbf{$\Delta_{acc}$}  & \textbf{Remain. (\%)} & \textbf{Bits} & \textbf{Comp.} \\
\textbf{} & \textbf{} & \textbf{}  & \textbf{acc.(\%)} & \textbf{acc.  (\%)}  & \textbf{}  & \textbf{params.} & \textbf{} & \textbf{rate} \\\midrule

 
 \textbf{CIFAR-10} & \textbf{} & \textbf{} & \textbf{} & \textbf{} & \textbf{} \\

\textbf{} &  \textbf{VGG-16} & \textbf{\cite{li2020penni}}      & 93.49 & 93.14 &  -0.35 & 55.56 & 32 & 1.80x\\
\textbf{} &  \textbf{} & \textbf{DPP-M} (This work)  & 93.50 & 93.60 & \textbf{+0.10} & 10.73 & 32 & 9.32x\\
\textbf{} &  \textbf{} & \textbf{DPP-M} (This work)  & 93.50 & 93.52 & +0.02 & 15.3  & 8  & \textbf{26.14x}\\

\textbf{} &             & \textbf{\cite{Zhao_2019_CVPR}}  & 93.25 & 93.18 & -0.07 & 26.66	& 32 & 3.75x\\
\textbf{} &  \textbf{} & \textbf{DPP-C} (This work)  & 93.50 & 93.52 & \textbf{+0.02}	& 19.49	& 32 & 5.13x\\
\textbf{} &  \textbf{} & \textbf{DPP-C} (This work)  & 93.50 & 93.10 & -0.4	& 15.61	& 8	 & \textbf{25.62x}\\ \midrule

\textbf{} & \textbf{MobileNet v1} & \textbf{\cite{lubana2020gradient}}        & 92.3 &	91.77	& -0.53	& 25 & 32 & \textbf{4x} \\
\textbf{} &                       &\textbf{DPP-C} (This work) & 93.6 &	93.14	& \textbf{-0.46}& 36,95	& 32 & 2.70x\\ \midrule

 \textbf{CIFAR-100} & \textbf{} & \textbf{} & \textbf{} & \textbf{} & \textbf{} \\
\textbf{} &  \textbf{VGG-16} & \textbf{\cite{Zhao_2019_CVPR}}  & 73.26 & 73.33  & +0.07  & 68 & 32 & 1.47x \\
    \textbf{} & \textbf{}        & \textbf{DPP-C} (This work)       & 70.32 & 70.40 &  \textbf{+0.08} & 18.8 & 32 & \textbf{5.31x}  \\ \midrule

\textbf{} & \textbf{MobileNet v1} & \textbf{\cite{lubana2020gradient}} & 69.1 &	68.52	& -0.58	& 35 & 32 & \textbf{2.85x} \\
\textbf{} &                       &\textbf{DPP-C} (This work) & 72.35 &	72.50	& \textbf{-0.15} & 40 & 32 & 2.50x\\

\bottomrule
\end{tabular}
\end{small}
\end{minipage}
\end{table*}

\textbf{Training details.} For training both networks we use Adam optimizer with a learning rate of 0.001, $\mu$ is set to 0.005, while $\beta$ is set to $1$ . We used a batch size of 128, and all weights are initialized with Xavier uniform initialization 

\subsection{CIFAR-10 and CIFAR-100}

We evaluate DPP-C and DPP-M on VGG-16 \cite{Simonyan2015VeryDC} and MobileNet v1 for both CIFAR-10 and CIFAR-100. For VGG-16 trained on CIFAR-10, we evaluate DPP-M with PENNI-D. PENNI-D was recently proposed by \cite{li2020penni}, and similarly provides a method for kernel pruning. Notice that \cite{li2020penni} provides a four-stage pruning pipeline, therefore we based our comparison with PENNI-D (without the additional shrinkage stage). DPP-C is compared with the recent filter pruning algorithm of \cite{Zhao_2019_CVPR} for both cases, CIFAR-10 and CIFAR-100. Finally, DPP-C is compared with \cite{lubana2020gradient} that performs filter pruning. Relevantly, for all fully-connected layers, DPP-F is applied, therefore the networks are completely pruned in all stages (convolutional and fully-connected layers), in comparison with most of conventional structured pruning approaches. Results are shown in Table~\ref{table:2}.

\textbf{Training details.} Training details for CIFAR-10 and CIFAR-100 include a learning rate schedule, which halves the learning rate every 40 epochs. $\mu$ is set to 0.005, while $\beta$ is set to $0.1$ SGD optimizer is used with momentum=0.9. Data augmentation is used for these experiments. All weights are initialized using He normal initialization 

\subsection{ImageNet}

We test DPP-C on ResNet18 trained on the ImageNet dataset \cite{Krizhevsky12}. 
We train the model with an initial learning rate of 0.1, which we decreased by a factor of 10 after 30 epochs. We use the SGD optimizer with a momentum of 0.9. After training for 35 epochs, the top-1 test accuracy with 48.24\% of the filters dynamically pruned is 62.3\%, which is only 0.2\% lower than its non-pruned counterpart. 

\begin{figure}[h!]
     \centering
     \begin{subfigure}[]{}
         \centering
         \includegraphics[trim=0 15 0 10,width=0.43\columnwidth]{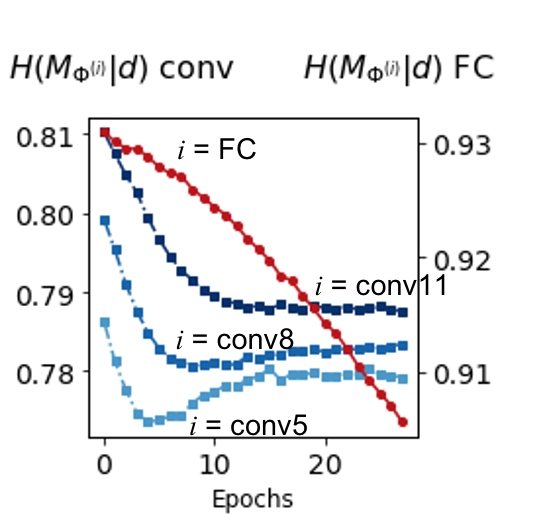}
         \label{fig:entropyvgg1}
     \end{subfigure} \quad
     \hfill
     \begin{subfigure}[]{}
         \centering
         \includegraphics[trim=0 15 0 10,width=0.43\columnwidth]{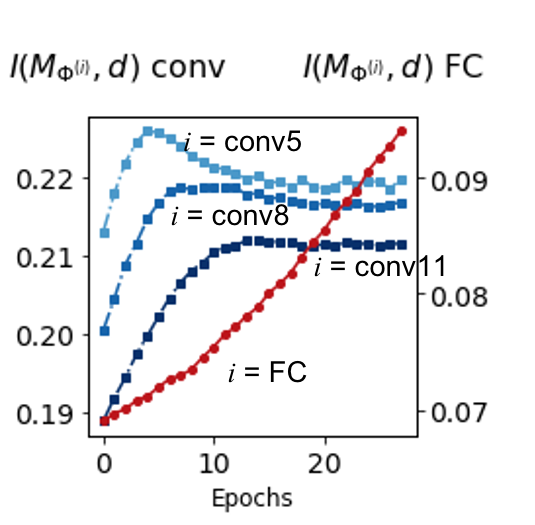}
         \label{fig:entropyvgg2}
     \end{subfigure} %

     \begin{subfigure}[]{}
         \centering
         \includegraphics[width=0.7\columnwidth]{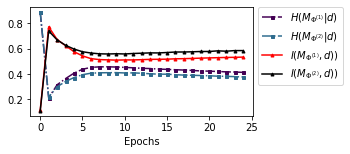}
         \label{fig:entropylenet}
     \end{subfigure} %
\caption{$H(\mask|d)$ and $I(\mask,d)$ for VGG16 (CIFAR10) are shown in (a) and (b) respectively. The plots correspond to $5^{th}$, $8^{th}$, and $11^{th}$ convolutional layers, and the $1^{st}$ fully-connected layer (FC). $H(\mask|d)$ and $I(\mask,d)$ for the first two fully-connected layers of LeNet300-100 are shown in (c).}
 \label{fig:entropy}
\end{figure}

\subsection{Analysis of sparsity belief over time}
To analyze the pruning behavior during training, we visualize the proposed metrics from Section \ref{sec:metrics} in Fig.~\ref{fig:entropy}, where we normalized by the upper bound of the entropy to facilitate straightforward comparison between layers. Recall that a low $\mathrm{Average~Pruning~Entropy} H(\mask|d)$ denotes a high confidence in pruning masks, whereas a high $\mathrm{Pruning~Diversity} I(\mask,d)$ indicates high mask diversity/specialization.

We find interesting dynamics in Fig. \ref{fig:entropy}a,b (VGG16 - CIFAR10), where convolutional layers seem to more quickly learn a diverse set of sparse patterns ($I(\mask,d)$ grows faster) with high confidence ($H(\mask|d)$ drops quicker) than fully-connected layers. 

Figure \ref{fig:entropy}c shows the aforementioned metrics for the first two layers of LeNet300-100. Both layers show similar behavior; a quick improvement both in confidence ($H(\mask|d)$) and diversity ($I(\mask,d)$) at the start of training, after which both metrics stabilize.

\section{Conclusion}
In this paper, we propose dynamic probabilistic pruning (DPP), an algorithm that enables training sparse networks based on stochastic and dynamic masking. DPP is a general framework that enables structured pruning for fully-connected and convolutional layers, suitable for hardware implementations. We demonstrated that DPP enables a larger hardware memory saving by leveraging structured pruning at different levels of granularities (fine, medium and coarse). Leveraging its probabilistic nature, we showed how one can assess the confidence and diversity of pruning masks among neurons by monitoring proposed information-theoretic metrics. 

Since DPP does not rely on magnitudes for determining the relevance of weights, it can be straightforwardly integrated with weight quantization (including binarization). This allows for a larger model compression as it is observed in the results. We test its performance for three benchmark datasets and obtain competitive accuracies for different architectures. In conclusion, our method generates ultra-compressed models, allowing the integration of quantization (and even binarization) and pruning, while providing a level of structured sparsity, enabling a more efficient implementation on existing hardware platforms. Further, the potential of DPP should be explored to generate even more efficient sparsity patterns for hardware such as tiling at different levels of granularities.


%





\ifCLASSOPTIONcaptionsoff
  \newpage
\fi



%



\bibliographystyle{ieeetr}
\bibliography{bibliography.bib}

\end{document}